\documentclass{article}

\usepackage{arxiv}

\usepackage[utf8]{inputenc} 
\usepackage[T1]{fontenc}    
\usepackage{hyperref}       
\usepackage{url}            
\usepackage{booktabs}       
\usepackage{amsfonts}       
\usepackage{nicefrac}       
\usepackage{microtype}      
\usepackage{lipsum}		
\usepackage{graphicx}
\usepackage{natbib}
\usepackage{doi}
\usepackage{xcolor}
\usepackage{bbding}
\usepackage{fourier}
\usepackage{tikz}

\title{Can Large Language Models Detect Methodological Flaws? Evidence from ‘Gesture Recognition for UAV-Based Rescue Operation Based on Deep Learning’}

\author{
 Domonkos Varga \\
}

\renewcommand{\shorttitle}{\textit{arXiv} Template}

\hypersetup{
pdftitle={A template for the arxiv style},
pdfsubject={q-bio.NC, q-bio.QM},
pdfauthor={Domonkos Varga},
pdfkeywords={Gesture recognition, Machine learning integrity, Large language models},
}

\begin{document}
\maketitle

\begin{abstract}
Reliable evaluation is a fundamental requirement in machine learning research, yet methodological 
flaws—particularly data leakage—continue to undermine the validity of reported results.
In this work, we investigate whether large language models (LLMs) can act as independent
analytical agents capable of identifying such flaws in published studies.
As a case study, we examine the paper “Gesture Recognition for UAV-Based Rescue Operation
Based on Deep Learning,” \cite{liu2021gesture}
which reports near-perfect classification accuracy on a small, human-centered dataset.
We first conduct a methodological analysis of the evaluation protocol and show that the
reported results are consistent with subject-level data leakage arising from non-independent
training and test splits. We then assess whether this issue can be detected independently
by multiple state-of-the-art LLMs. The original paper is provided to six models without
any prior context or critique, using an identical prompt that instructs them to evaluate
the methodology.
Across all models, we observe a striking level of agreement. Each LLM identifies
the evaluation protocol as flawed and attributes the reported performance to non-independent
data partitioning, supported by characteristic indicators such as near-identical learning
curves, minimal generalization gap, and an almost perfectly diagonal confusion matrix.
Despite differences in reasoning style and level of detail, the conclusions are consistent
across models.
These findings suggest that modern LLMs are capable of detecting common methodological
flaws in machine learning research based solely on published artifacts. While their
outputs should not be considered definitive proof, their consistent agreement highlights
their potential as auxiliary tools for improving reproducibility and supporting scientific
auditing. More broadly, this work contributes to the emerging paradigm of AI-assisted
meta-science, where automated systems complement human expertise in evaluating the
reliability of scientific results.
\end{abstract}


\section{Introduction}
\label{sec:intro}
Reliable evaluation is a cornerstone of machine learning research. Reported performance
metrics are only meaningful if the underlying experimental protocol ensures a strict
separation between training and test data. Violations of this principle --- commonly referred
to as \textit{data leakage} --- can lead to overly optimistic results that do not
reflect real-world
generalization. Such issues are particularly prevalent in human-centered recognition tasks,
where repeated measurements from the same individuals can inadvertently appear in both
training and test sets, resulting in \textit{subject leakage}. Despite extensive awareness of
these risks, methodological flaws in evaluation protocols continue to appear in the
literature, often remaining undetected during peer review.

At the same time, recent advances in large language models (LLMs) have demonstrated
strong capabilities in reasoning about code, experimental design, and statistical
patterns. Trained on vast corpora of scientific and technical text, modern LLMs exhibit
an emerging ability to identify inconsistencies, anomalies, and methodological weaknesses
in machine learning workflows. This raises an important question: can LLMs serve as
independent analytical agents capable of detecting methodological flaws --- such as data
leakage --- in published research?

In this work, we investigate this question through a focused case study of the
paper “Gesture Recognition for UAV-Based Rescue Operation Based on Deep Learning”
\cite{liu2021gesture}. The selected study reports near-perfect classification accuracy
(~99\%) on a gesture-recognition task using a dataset collected from a small number of 
participants. Such performance, while impressive at face value, is atypical for
real-world human action recognition and warrants careful methodological scrutiny.
The key bibliographic details of the target paper are
presented in Table \ref{tab:paper}, listing its title, authors,
publication venue, and year of appearance. This information
serves as a reference point for the critical examination
that follows.

\begin{table}
\caption{Reference Details of the Original Work \cite{liu2021gesture} Under Analysis.}
\centering
\begin{tabular}{ll}
\toprule
Title & Gesture Recognition for UAV-based Rescue Operation based on Deep Learning   \\
Authors & Chang Liu; Tam{\'a}s Szir{\'a}nyi \\
Book Title & Improve \\
Pages & 180--187 \\
Publication Year & 2021 \\
\bottomrule
\end{tabular}
\label{tab:paper}
\end{table}

To assess whether LLMs can independently identify potential flaws, we submitted the full paper
as a PDF to six state-of-the-art models—GPT-5.2, Claude Sonnet 4.6, Google Gemini 3.0 Pro
(Thinking mode), Kimi 2.5 (Instant), DeepSeek-V3, and GLM-5—using an identical,
expert-level prompt. Each model was asked to analyze the methodology with a specific focus
on the evaluation
protocol, confusion matrix, and learning curves, and to determine whether the reported results
are free from data leakage.
Importantly, the models were queried independently, without any prior context or
knowledge of existing critiques. This setup enables a form of consensus-based validation:
if multiple, architecturally distinct LLMs converge on the same diagnosis, it provides
evidence that the underlying methodological issue is detectable from the published material
alone, rather than requiring external domain knowledge or subjective interpretation.

The results show a striking level of agreement. Across all six models, the conclusions
consistently indicate the presence of non-independent training and test splits, most
plausibly arising from frame-level random partitioning of data collected from the same
subjects. The models further identify characteristic warning signs of data leakage,
including near-perfect confusion matrices, highly synchronized training and testing
curves, and test performance that matches or exceeds training performance throughout
the learning process.

\subsection{Contributions}
\label{sec:contri}
The contribution of this paper is twofold. First, it provides additional evidence—independent
of prior human analysis—that the evaluation protocol in the examined study does not support
claims of generalization to unseen individuals. Second, it demonstrates that modern LLMs
can function as \textit{auxiliary methodological auditors}, capable of detecting
common evaluation flaws
in machine learning research. While LLM outputs should not be treated as authoritative proof,
their consistent agreement suggests that they can play a valuable role in
supporting reproducibility and improving the robustness of scientific evaluation.

Overall, this work contributes to the emerging area of \textit{AI-assisted scientific
auditing}, highlighting both the potential and the limitations of using LLMs as tools for identifying methodological weaknesses in published research.

\subsection{Structure of the paper}
\label{sec:structure}
The remainder of this paper is organized as follows.
Section \ref{sec:related} reviews the relevant literature on data leakage in machine learning and
gesture recognition.
Section \ref{sec:summary_1} describes the methodology used for the LLM-based analysis.
Section \ref{sec:leakage} provides a detailed examination of the evaluation protocol and the identified
sources of data leakage. Section \ref{sec:llm}
presents the independent assessments produced by multiple large language models and
analyzes their consistency. Finally, Section \ref{sec:conc} concludes the paper.

\section{Related work}
\label{sec:related}
This section reviews the two main areas of research relevant to the present study. First,
we summarize prior work on data leakage in machine learning, with a particular focus on it
s impact on evaluation validity and reproducibility. Second, we provide an overview of
gesture-recognition methods, emphasizing the characteristics of human-centered datasets
and the importance of subject-independent evaluation. Together, these perspectives establish
the methodological context in which the analyzed study \cite{liu2021gesture} is
situated and motivate the
need for rigorous evaluation protocols.
\subsection{Data leakage in machine learning research}
\label{sec:leakage}
Data leakage is a well-documented yet persistently recurring issue in machine 
learning research, arising when information from outside the intended
training distribution is inadvertently incorporated into the model during
training or evaluation. This violates the fundamental assumption of
independence between training and test data, leading to overly optimistic
performance estimates that do not reflect true generalization
capabilities \cite{domnik2022data}, \cite{apicella2025don}.

Several forms of data leakage have been identified in the literature.
These include improper preprocessing applied jointly to training and test
data, duplication of samples across partitions, and the use of features
that implicitly encode target information. In sequential settings, temporal
leakage may occur when future observations influence past predictions. In
human-centered recognition tasks—such as gesture or action recognition—subject
leakage represents a particularly critical case, where the same individuals
appear in both training and test sets. Because samples originating from the same
subject share strong correlations (e.g., body proportions, habitual motion patterns,
or execution style), models may learn subject-specific cues instead of the
underlying task-relevant structure.

The broader implications of data leakage have been extensively discussed in
recent meta-scientific analyses. \cite{kapoor2023leakage}, for
example, demonstrated that leakage-related issues contribute significantly
to the reproducibility crisis in machine-learning-based science. In their survey
of multiple disciplines, they identified hundreds of studies affected by various
forms of leakage and proposed a taxonomy of common failure modes. These include
lack of a clean separation between training and test data (L1), duplicate samples
across datasets (L1.4), and non-independence between training and test
observations (L3.2), among others. Such findings highlight that leakage is not
merely a technical oversight, but a systemic challenge that can undermine the validity
of reported results across fields.

Detecting data leakage is often non-trivial, as it does not necessarily produce explicit errors
during model training. Instead, it manifests indirectly through characteristic patterns in
evaluation metrics. Models affected by leakage frequently exhibit unusually high accuracy,
near-perfect confusion matrices, and training and testing curves that evolve in close synchrony.
In some cases, test performance may match or even exceed training performance, indicating that
the test set does not represent a truly independent distribution. These patterns, while not
definitive proof on their own, are widely recognized as strong warning signs of methodological
flaws in the evaluation protocol.

Despite increased awareness and established best practices, leakage-related issues continue to
appear in published work. One contributing factor is the growing complexity of machine learning
pipelines, where multiple preprocessing and feature extraction steps increase the risk of
unintended information flow. Another factor is the limited transparency in reporting experimental
procedures, which makes it difficult for reviewers and readers to fully assess whether proper
data separation has been maintained.

To address these challenges, standard methodological guidelines emphasize strict separation of
training and test data, alignment of evaluation protocols with the intended deployment scenario,
and explicit documentation of data partitioning strategies. In domains involving repeated 
measurements from the same entities—such as human subjects—this typically requires partitioning 
data at the entity level rather than at the sample level. Only such subject-independent evaluation
protocols can provide a reliable estimate of a model’s ability to generalize to previously unseen
individuals.

In this context, the present work extends prior methodological analyses by exploring whether LLMs
can independently identify characteristic indicators of data leakage based solely
on reported experimental artifacts. By situating this investigation within the broader literature
on leakage and reproducibility, we aim to contribute to ongoing efforts toward more robust and 
transparent evaluation practices in machine learning research.

\subsection{Gesture recognition}
\label{sec:gesture}
Human gesture recognition has been extensively studied across multiple
disciplines, including computer vision, robotics, and human–machine
interaction \cite{al2020review}, \cite{dallel2020inhard}, \cite{gammulle2023continuous}.
The goal of gesture recognition systems is to interpret human body movements or poses
and map them to predefined semantic categories, enabling natural communication
between humans and intelligent systems. In recent years, the field has gained particular
attention in applications such as surveillance, assistive technologies, and autonomous
systems, including UAV-based human–machine interaction
scenarios \cite{ma2017hand}, \cite{perera2018uav}.

Gesture recognition can be approached using a variety of sensing modalities,
including wearable inertial sensors \cite{jiang2021emerging},
electromyography (EMG) \cite{jaramillo2020real}, depth cameras \cite{suarez2012hand},
radar-based systems \cite{ahmed2021hand}, and multimodal fusion 
approaches \cite{liu2023multimodal}.
Among these, vision-based methods remain one of the most widely adopted
paradigms due to their non-intrusive nature and applicability in remote
or unconstrained environments. In such systems, gestures are typically
inferred from RGB video, optical flow, or higher-level representations
derived from visual input, such as human pose or skeletal models
\cite{dos2020dynamic}, \cite{holte2010view}, \cite{zhou2016novel}.

An alternative and increasingly popular approach is pose-based gesture
recognition, where human skeletal keypoints are extracted using frameworks such as
OpenPose or MediaPipe \cite{viswakumar2019human}, \cite{lugaresi2019mediapipe}.
In these systems, the pose estimator serves as a front-end that converts raw images into
structured representations of joint coordinates, which are then processed by downstream 
classifiers. Skeleton-based representations offer several advantages, including reduced 
dimensionality, robustness to background variation, and partial invariance to
appearance-related factors such as clothing or lighting. These properties make pose-based
methods particularly suitable for applications involving aerial platforms, where viewpoint
and environmental conditions can vary significantly.

Despite these advances, a recurring challenge in gesture-recognition research is
the limited diversity and scale of many available datasets \cite{saupe2016crowd}.
Gesture execution varies substantially across individuals due to differences in body
proportions, movement styles, and contextual factors. As a result, evaluation protocols
have increasingly emphasized subject-independent validation, in which models are tested on
individuals not seen during training. This approach is essential for assessing real-world
generalization, especially in applications such as UAV-based rescue, where systems must
interpret gestures from previously unseen users.
Nevertheless, not all studies adhere to these best practices. In particular, the use of
sample-level or frame-level random splits in datasets containing repeated measurements
from the same individuals can lead to unintended subject overlap between training and
test sets. Such evaluation strategies may yield artificially high accuracy while failing
to measure true gesture-recognition capability. This highlights the importance of carefully
designed experimental protocols that align with the intended deployment scenario and ensure
the validity of reported performance metrics.

Despite the extensive body of work on gesture recognition and the well-established awareness of
data leakage in machine learning, the intersection of these two areas remains insufficiently
explored. In particular, while subject-independent evaluation is widely recognized as essential
in human-centered recognition tasks, deviations from this principle continue to appear in
practice, often accompanied by unrealistically high reported performance.

At the same time, recent advances in large language models raise the question of whether such
methodological issues can be detected not only through expert analysis, but also via independent,
automated reasoning. If characteristic indicators of data leakage—such as anomalous learning 
curves or near-perfect confusion matrices—are sufficiently explicit in published results, then
both human experts and modern LLMs may be able to identify them without additional context.

Motivated by this observation, the present work investigates a concrete case in
gesture-recognition research and examines whether multiple independent analytical agents—both
human and LLM-based—converge on the same methodological diagnosis. In doing so, we aim to
bridge the gap between established knowledge on evaluation pitfalls and emerging approaches
to automated scientific auditing.

\section{Summary of the gesture-recognition method by Liu \& Szir\'{a}nyi}
\label{sec:summary_1}
\begin{figure}
\includegraphics[width=0.75\textwidth]{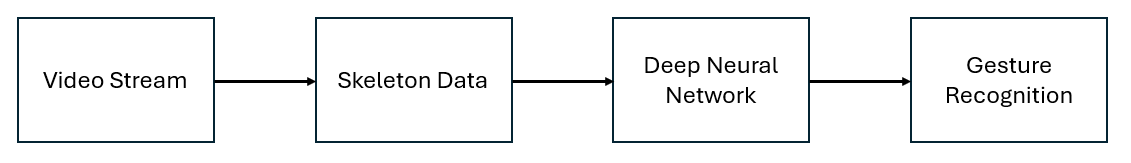}
\caption{Workflow of the gesture-recognition system implemented and proposed by \cite{liu2021gesture}.}
\label{fig:method}
\end{figure}

The method proposed by \cite{liu2021gesture} presents a vision-based
gesture-recognition
system designed to support human–UAV interaction in rescue scenarios. The overall
pipeline integrates human detection, pose estimation, tracking, and classification into
a unified framework operating on aerial video data (depicted in Figure \ref{fig:method}).

At the core of the approach is a pose-based representation of human motion. For each
frame of the input video stream, human skeletal keypoints are extracted using the
OpenPose \cite{qiao2017real} framework, which provides 2D coordinates
of 18 body joints per detected person.
These skeletal features serve as the primary input for gesture recognition, offering a compact
and interpretable representation of human posture while reducing sensitivity to background
variation. In addition to pose extraction, the system incorporates multi-person tracking 
using the Deep SORT \cite{wojke2017simple} algorithm. This component enables consistent 
identity assignment
across frames and allows the system to handle scenarios involving multiple individuals
simultaneously. Based on the detected persons, the system distinguishes between
different interaction cases (e.g., no person detected, single individual, or 
multiple individuals) and activates gesture recognition accordingly.

The gesture-recognition task itself is formulated as a supervised classification
problem over a predefined set of ten body gestures, including actions such as
\textit{Kick}, \textit{Punch}, \textit{Squat}, \textit{Stand}, \textit{Walk}, and \textit{Sit},
as well as two application-specific dynamic
gestures: \textit{Attention} and \textit{Cancel}. These latter gestures are designed to
initiate and
terminate interaction with the UAV, functioning as control signals in rescue scenarios.
To construct the dataset, the authors collected aerial video recordings using a UAV-mounted
camera. The dataset consists of gesture performances from six participants, each executing
multiple variations of the defined gesture set. From these recordings, skeletal data were
extracted frame-by-frame and used to form the training samples.

For classification, the authors evaluated several machine learning models, including
k-nearest neighbors (kNN), support vector machines (SVM), random forests, and a deep
neural network (DNN). Based on empirical performance, the DNN was selected as the final
model. The network is implemented as a fully connected feed-forward architecture with
four dense layers (128–64–16–10 units), batch normalization, and a Softmax output layer
corresponding to the ten gesture classes. Training is performed using categorical 
cross-entropy loss and the Adam optimizer \cite{adam2014method}.

The dataset is partitioned into training and test sets using a 90\%–10\% split.
The model is trained for 50 epochs, and performance is evaluated using accuracy,
loss curves, confusion matrices, and macro-F1 scores. The reported results indicate
near-perfect classification performance, with approximately 99\% accuracy on both
training and test sets, and a confusion matrix that is largely diagonal.

\section{Detected data leakage in the evaluation protocol of Liu \& Szir\'{a}nyi}
\label{sec:leakage}
The evaluation protocol reported by \cite{liu2021gesture} raises fundamental
concerns regarding the independence of the training and test sets. In this
section, we analyze the data partitioning strategy and demonstrate that the
reported results are consistent with a non-independent evaluation setup, most
plausibly affected by subject-level data leakage.

A key observation follows directly from the dataset description and the applied
splitting procedure. The dataset was collected from six individuals, and the full
set of samples was divided into training and test subsets using a 90\%–10\% split.
Under such conditions, a subject-independent partitioning is not feasible if the split
is performed at the sample level. Any random division of individual frames or samples
will necessarily distribute data from each participant across both the training and test
sets. Consequently, the same individuals appear in both partitions, violating the
independence assumption required for valid generalization assessment.

\begin{figure}
\includegraphics[width=0.95\textwidth]{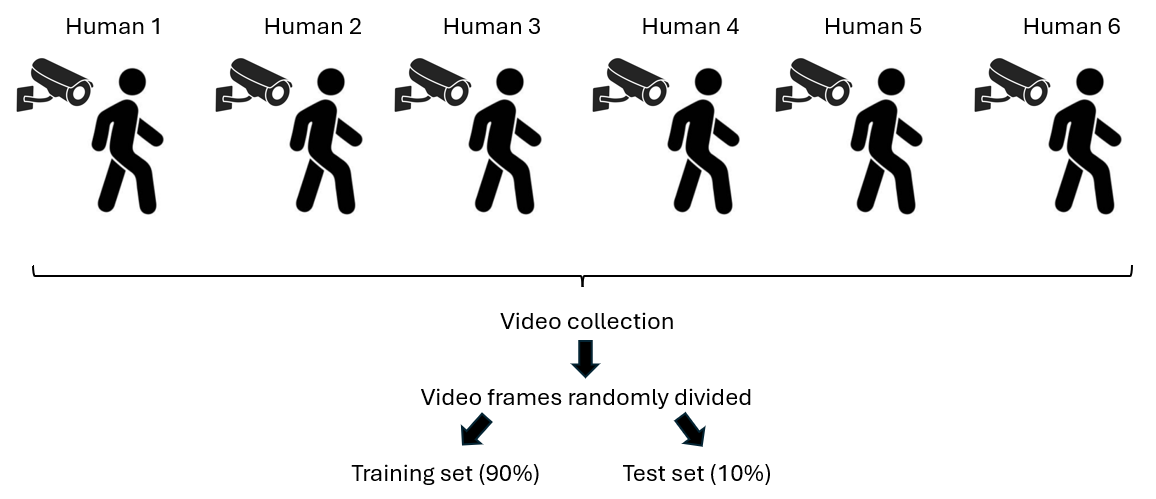}
\caption{Illustration of the likely incorrect data-splitting
procedure used in the original study. Video recordings from all
six participants appear to have been merged into a single
pool, from which individual frames were randomly divided into a
90\% training set and a 10\% test set. This frame-level random 
split necessarily mixes data from each person into both sets,
resulting in unavoidable subject leakage.}
\label{fig:incorrect}
\end{figure}

This issue is illustrated conceptually in Figure \ref{fig:incorrect}, which depicts the likely
data-splitting strategy implied by the original description. In this scenario,
all recorded data are first aggregated into a single pool, after which individual samples
(e.g., frames or extracted skeletal representations) are randomly assigned to the training
and test sets. Because the samples originate from a small, fixed set of subjects, this
procedure results in substantial overlap between the two partitions at the subject level.
As a consequence, the model is evaluated on data that are statistically highly similar to 
those seen during training—not only in terms of gesture dynamics, but also with respect to
subject-specific characteristics such as body proportions, habitual motion patterns, and execution style.

\begin{figure}
\includegraphics[width=0.95\textwidth]{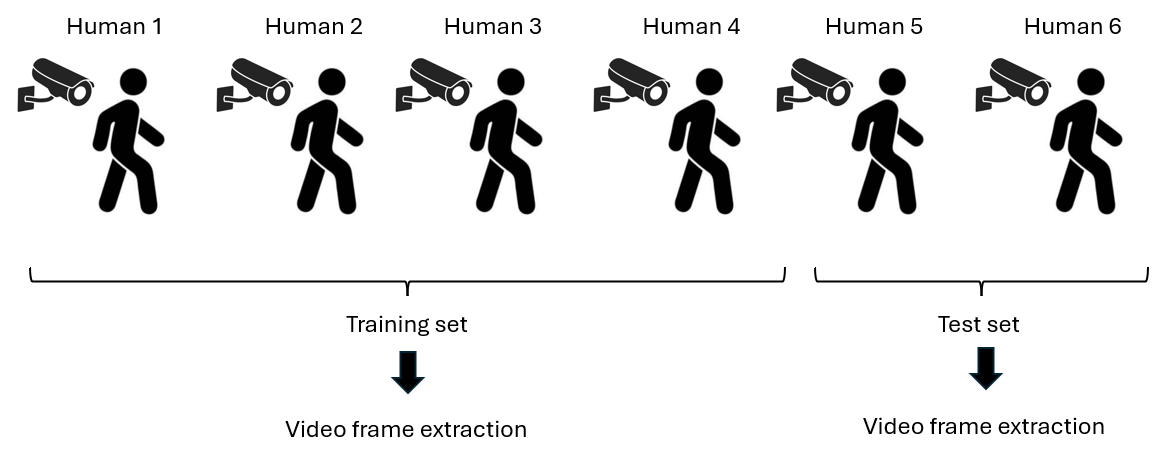}
\caption{Correct subject-independent data-splitting protocol. Entire participants are assigned exclusively to the training or test set before any video frames are extracted. This prevents subject overlap and ensures that the test set contains individuals unseen during training, enabling a valid evaluation of generalization.}
\label{fig:correct}
\end{figure}

In contrast, Figure \ref{fig:correct} illustrates the subject-independent evaluation protocol that would
be required to properly assess generalization. In this correct setup, entire subjects are 
assigned exclusively to either the training or the test set prior to any sample extraction.
All frames or derived features from a given individual are therefore confined to a single 
partition, ensuring that the model is evaluated on previously unseen subjects. This form
of partitioning is widely recognized as standard practice in human action and gesture
recognition tasks, where intra-subject correlations are strong and can otherwise bias
performance estimates.

\begin{figure}
\includegraphics[width=0.55\textwidth]{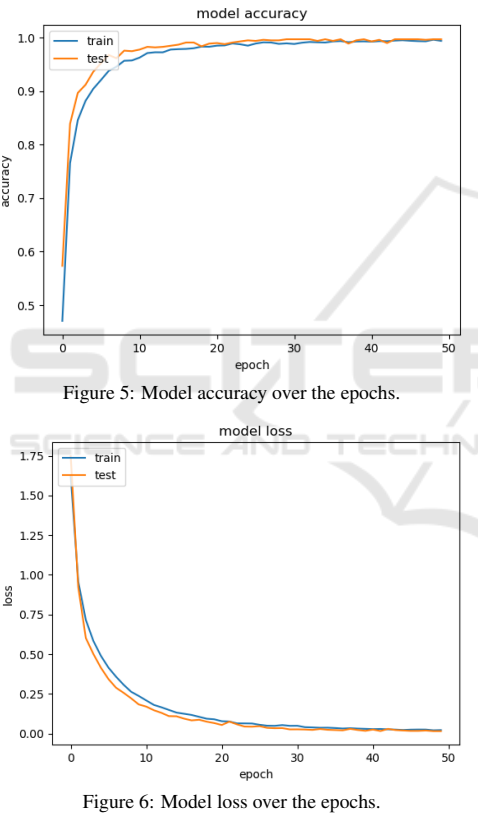}
\caption{The training curves published in \cite{liu2021gesture}.}
\label{fig:accuracy}
\end{figure}

Beyond the structural argument, several empirical indicators reported in the original 
study further support the presence of non-independent data partitions. The published 
learning curves (shown in Figure \ref{fig:accuracy}) show near-identical trajectories
for training and test accuracy,
converging
rapidly toward near-perfect performance. Similarly, the corresponding loss curves decrease
in close synchrony, with minimal divergence throughout training. In realistic settings 
involving human variability, such behavior is uncommon; test performance typically 
exhibits greater variability and lags behind training performance due to differences
between seen and unseen data.

\begin{figure}
\includegraphics[width=0.75\textwidth]{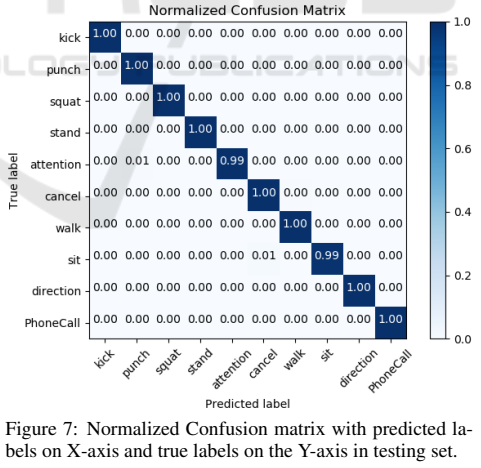}
\caption{The normalized confusion matrix published
in \cite{liu2021gesture}.}
\label{fig:confusion}
\end{figure}

Additional evidence is provided by the reported confusion
matrix (shown in Figure \ref{fig:confusion}), which is almost perfectly
diagonal, indicating near-zero misclassification across all gesture classes. While high
accuracy is not inherently problematic, the combination of near-perfect classification, 
minimal train–test divergence, and a small, homogeneous dataset strongly suggests that
the model is benefiting from shared information between the training and test sets rather
than learning generalizable gesture representations.

Taken together, the dataset composition, the splitting strategy, and the
reported performance characteristics form a consistent picture: the evaluation
protocol does not enforce independence between training and test data. As a result,
the reported metrics are likely to overestimate the true generalization capability of
the model, particularly in scenarios involving previously unseen individuals. This limitation
is especially critical in the context of UAV-based rescue applications, where reliable
performance on unknown users is a fundamental requirement.

Importantly, this analysis does not question the implementation of the proposed model
itself, but rather the validity of the evaluation procedure used to assess it. The
findings highlight the necessity of subject-independent data partitioning in human-centered
machine learning tasks and underscore how seemingly strong quantitative results can arise
from methodological artifacts when evaluation protocols are not carefully designed.

\begin{figure}
\includegraphics[width=0.95\textwidth]{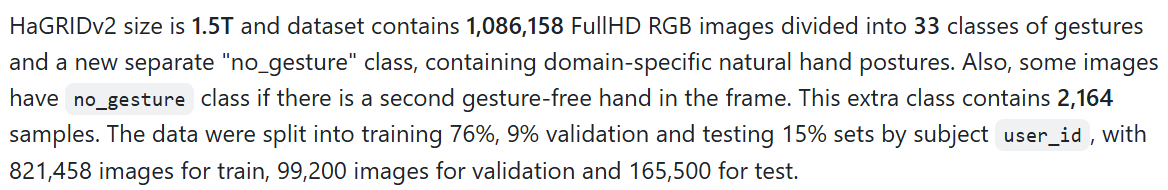}
\caption{Excerpt from the official HaGRID benchmark
(available at: \url{https://github.com/hukenovs/hagrid})
documentation illustrating its subject-independent
data-splitting protocol. The dataset is partitioned
by \textit{user\_id} into 76\% training, 9\% validation,
and 15\% testing, ensuring that each subject appears
exclusively in a single split. This exemplifies current
best practice in gesture-recognition research and contrasts
sharply with frame-level random splitting applied by \cite{liu2021gesture}.}
\label{fig:hagrid}
\end{figure}

A further argument in favor of subject-independent evaluation can be found in recent
large-scale gesture-recognition benchmarks. For instance, the HaGRID
dataset \cite{kapitanov2024hagrid}, a modern and high-quality benchmark developed specifically
for hand-gesture recognition, explicitly adopts a subject-wise partitioning
strategy (Figure \ref{fig:hagrid}). In this setup, the data are divided
into 76\% training, 9\% validation, and 15\% testing sets based on user identifiers,
ensuring that each individual appears exclusively in a single split.
This design choice reflects current best practices in gesture-recognition research,
where reliable evaluation requires the separation of entire subjects rather than the
random distribution of their samples across partitions. The fact that contemporary
benchmark datasets enforce subject-level isolation highlights the importance of this
principle and underscores that frame-level random splitting strategies—such as the one
applied in the analyzed study—are not suitable for assessing generalization to
previously unseen individuals.

\section{Independent LLM-Based Assessment of the Evaluation Protocol}
\label{sec:llm}
To further examine whether the methodological issues identified in the previous section
can be detected independently, we conducted an additional analysis using multiple
LLMs. The objective of this experiment was not to treat LLM outputs as authoritative
evidence, but rather to assess whether modern models --— when prompted as domain
experts -—- can identify potential data leakage and evaluation flaws based solely on
the information contained in the original publication.
Importantly, each LLM was queried in isolation, without any prior context,
explanation, or reference to the critique presented in this paper. The original
article by \cite{liu2021gesture} was provided as a standalone
PDF document, and the models were instructed to analyze the methodology from first
principles. This setup ensures that the responses reflect independent reasoning based
only on the reported experimental design, results, and visualized metrics.

All models were prompted using the same instruction:
\begin{quote}
\textit{
Assume you are a data scientist. Analyze the methodology described in the provided PDF
and determine whether the procedure is free from data leakage. Examine in particular
the evaluation protocol, the published confusion matrix, and the learning curves. 
Identify any potential sources of leakage or methodological flaws and justify your assessment.
}
\end{quote}
The following subsections publish the responses generated by six state-of-the-art
LLMs: GPT-5.2, Claude Sonnet 4.6, Google Gemini 3.0 Pro (Thinking mode), Kimi 2.5
(Instant), DeepSeek-V3, and GLM-5.

Very important to underscore that several measures were taken to ensure the validity of the
analysis and to reduce potential sources of bias.
First, the evaluation was conducted in a blind setting, where no prior critique or
hypothesis was provided to the models. This prevents confirmation bias and ensures that
the outputs reflect independent reasoning based on the input data.
Second, all models were queried using an identical prompt and the same input document.
This ensures consistency across evaluations and eliminates variability arising from prompt design.
Third, no model-specific tuning, system prompts, or iterative interactions were used. This 
avoids bias introduced by prompt engineering and ensures that the results reflect
baseline model behavior.
Fourth, the analysis focuses on convergent evidence rather than individual model correctness.
LLM outputs are not treated as authoritative judgments, but as independent analytical 
perspectives. Consistency across multiple models is interpreted as an indicator that
the identified issue is robust and detectable.

\subsection{GPT-5.2}
Here is a critical data science assessment of the methodology in the paper,
focusing on data leakage, evaluation protocol, confusion matrix, and learning curves.
\newline
\textbf{1. Evaluation protocol — major red flags}
\newline
\textcolor{red}{\textbf{!}} \textbf{Random split on a highly correlated dataset}
\newline
The dataset:
\begin{itemize}
    \item Collected from only 6 subjects.
    \item With “all possible variations for all gestures” recorded from the same individuals.
    \item Then split 90\% train / 10\% test.
\end{itemize}
\textbf{Why this is problematic}
\newline
This almost certainly implies:
\begin{itemize}
    \item Frame-level or sample-level random splitting
    \item Same subjects (and likely same recording sessions) appear in both train and test
\end{itemize}
\HandRight This is a classic data leakage scenario via identity leakage / temporal correlation.
\newline
In vision tasks, especially with:
\begin{itemize}
    \item Same background
    \item Same camera
    \item Same subjects
    \item Same session
\end{itemize}
the model can learn:
\begin{itemize}
    \item Subject-specific skeleton patterns
    \item Camera/viewpoint biases
    \item Recording artifacts
\end{itemize}
instead of true gesture generalization.
\newline
\textbf{Consequence}
\newline
The reported:
\begin{itemize}
\item 99.47\% train accuracy
\item 99.09\% test accuracy
\end{itemize}
are not reliable indicators of real-world performance.
\newline
\textbf{2. Dataset design — structural leakage risk}
\newline
\textcolor{red}{\textbf{!}} \textbf{Extremely small and homogeneous dataset}
\begin{itemize}
    \item Only 6 people
    \item Same lab environment
    \item Simulated “field” but actually controlled setup
\end{itemize}
\textbf{Implicit leakage sources}
\newline
\begin{enumerate}
    \item Subject identity leakage
    \begin{itemize}
        \item Skeleton proportions (limb lengths, posture style) are consistent per person
        \item Model may classify \textit{who} instead of \textit{what gesture}
    \end{itemize}
    \item Background / scene leakage
    \begin{itemize}
        \item Same room, same lighting
        \item Cropped images (explicitly mentioned)
    \end{itemize}
    \item Session leakage
    \begin{itemize}
        \item If sequences are split randomly, adjacent frames go to both sets
    \end{itemize}
\end{enumerate}
\HandRight This leads to \textbf{over-optimistic evaluation}.
\newline
\textbf{3. Learning curves — suspiciously perfect convergence}
\newline
From Figures 5–6 (page 6):
\begin{itemize}
    \item Training and test accuracy:
    \begin{itemize}
        \item Rapid rise
        \item Near-perfect convergence ($\approx$99\%)
    \end{itemize}
    \item Loss curves:
    \begin{itemize}
        \item Smooth, monotonic decrease
        \item No divergence between train/test
    \end{itemize}
\end{itemize}
\textbf{Interpretation}
\newline
This pattern is \textbf{not typical} for:
\begin{itemize}
    \item Small datasets
    \item Real-world gesture variability
\end{itemize}
Instead, it suggests:
\begin{itemize}
    \item \textbf{Low task difficulty OR leakage}
    \item Strong similarity between train and test distributions
\end{itemize}
\HandRight In this context, the most plausible explanation is \textbf{data leakage / non-independent test set}.
\newline
\textbf{4. Confusion matrix — unrealistically clean}
\newline
From Figure 7 (page 6):
\begin{itemize}
    \item Nearly perfect diagonal
    \item Almost zero off-diagonal errors
\end{itemize}
\textbf{Why this is suspicious}
\newline
With:
\begin{itemize}
    \item 10 gesture classes
    \item Real human variability
\end{itemize}
You would expect:
\begin{itemize}
    \item Confusion between similar gestures (e.g., Stand vs Attention, Walk vs Direction)
\end{itemize}
But:
\begin{itemize}
    \item Errors are nearly nonexistent
\end{itemize}
\HandRight This strongly suggests:
\begin{itemize}
    \item Test samples are too similar to training samples
    \item Likely same subjects + same conditions
\end{itemize}
\textbf{5. Feature pipeline — indirect leakage channel}
\newline
Pipeline:
\begin{itemize}
    \item OpenPose $\rightarrow$ skeleton $\rightarrow$ classifier
\end{itemize}
Even though skeletons reduce raw pixel bias, leakage still occurs via:
\begin{itemize}
    \item Body proportions (identity features)
    \item Consistent pose execution style
    \item Viewpoint invariance due to fixed setup
\end{itemize}
\HandRight Skeleton-based systems are not immune to leakage.
\newline
\textbf{6. Missing safeguards (critical methodological flaws)}
\newline
The paper does not mention:
\newline
\textcolor{red}{\XSolid} \textbf{No subject-wise split}
\newline
Should have used:
\begin{itemize}
    \item Leave-one-subject-out (LOSO)
    \item Or subject-disjoint train/test
\end{itemize}
\textcolor{red}{\XSolid} \textbf{No session-wise split}
\newline
Should separate:
\begin{itemize}
\item Different recording sessions
\end{itemize}
\textcolor{red}{\XSolid} \textbf{No cross-validation}
\newline
Single 90/10 split $\rightarrow$ high variance, unreliable
\newline
\textcolor{red}{\XSolid} \textbf{No external validation}
\newline
No:
\begin{itemize}
    \item Different environments
    \item Different users
    \item Real UAV deployment evaluation
\end{itemize}
\textbf{7. Additional methodological issues}
\newline
\warning \textbf{Class imbalance}
\newline
Dataset counts vary significantly (e.g., Cancel = 1994 vs Punch = 583)
\newline
But:
\begin{itemize}
    \item No discussion of balancing
    \item No weighted metrics
\end{itemize}
\warning \textbf{Evaluation metric bias}
\newline
\begin{itemize}
    \item Focus on accuracy and macro-F1 (both near 1.0)
    \item No per-class breakdown beyond confusion matrix
\end{itemize}
\warning \textbf{Overfitting not properly assessed}
\newline
They conclude: "well-fitting model” based on convergence.
\newline
But:
\begin{itemize}
    \item Convergence $\neq$ generalization
    \item Especially under leakage
\end{itemize}
\textbf{8. Final verdict}
\newline
\textcolor{red}{\XSolid} \textbf{The procedure is NOT free from data leakage}
\newline
Strong evidence of leakage:
\begin{itemize}
    \item Random split on same subjects
    \item Tiny, homogeneous dataset
    \item Near-identical train/test performance
    \item Almost perfect confusion matrix
    \item Lack of subject/session separation
\end{itemize}
\textbf{9. What would fix the methodology}
\newline
To make it valid:
\newline
\textcolor{purple}{\checkmark} \textbf{Proper evaluation}
\begin{itemize}
    \item Subject-wise split (LOSO or group split)
    \item Session-wise separation
\end{itemize}
\textcolor{purple}{\checkmark} \textbf{Robust validation}
\begin{itemize}
    \item Cross-validation across subjects
    \item External test set (new people, new environment)
\end{itemize}
\textcolor{purple}{\checkmark} \textbf{Realistic metrics}
\begin{itemize}
    \item Per-class precision/recall
    \item Performance drop across domains
\end{itemize}
\textbf{Bottom line}
\newline
The reported $\approx$99\% accuracy is \textbf{very likely inflated} due to \textbf{data
leakage and non-independent test data}. The methodology does not support claims
of real-world generalization for UAV rescue scenarios.
\subsection{Claude Sonnet 4.6}

\begin{figure}
\includegraphics[width=0.75\textwidth]{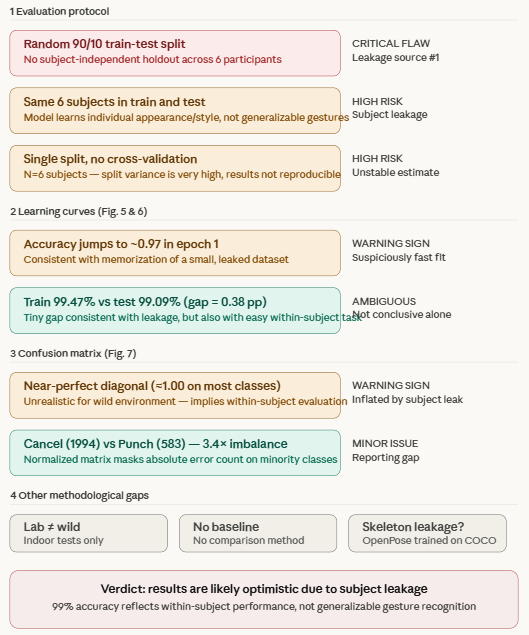}
\caption{Image generated by Claude Sonnet 4.6.}
\label{fig:claude}
\end{figure}

Overall verdict: the reported 99\% accuracy is almost certainly inflated, and the primary cause is subject leakage — a form of data leakage specific to person-centric recognition tasks.

\textbf{The core problem: no subject-independent split}

The most serious flaw is the evaluation protocol. The paper uses a single random 90/10 split over frames (or skeleton samples) drawn from a dataset collected from only 6 people. This means the same person's gestures appear in both the training and test set. The model can therefore learn individual body proportions, movement style, and even clothing artifacts rather than generalizable gesture patterns. When test samples come from the same people the model trained on, near-perfect accuracy is unsurprising — and says almost nothing about how the system would perform on a new person in the field. The correct approach would be leave-one-subject-out cross-validation (LOSO-CV), where each fold trains on 5 subjects and tests on the held-out 6th.

\textbf{What the learning curves tell us}

The accuracy curve (Fig. 5) shoots to roughly 0.97 in epoch 1 and converges smoothly by epoch 30. This is a red flag. Genuinely difficult generalization problems — the kind you encounter in wild rescue environments — show slower, noisier initial learning and larger epoch-to-epoch variance. Instant, smooth convergence is consistent with the model exploiting memorized within-subject patterns rather than learning robust features.
The 0.38 percentage point gap between training (99.47\%) and test (99.09\%) is unusually small, but is not conclusive on its own. A legitimate reason for a small gap would be a very easy task with abundant data — which is precisely what within-subject evaluation creates.

\textbf{What the confusion matrix tells us}

A near-perfect diagonal across all 10 classes under realistic outdoor conditions would be extraordinary. The literature on cross-subject action and gesture recognition typically reports accuracies in the 70–90\% range even on cleaner lab setups. The confusion matrix here is entirely consistent with a within-subject evaluation scenario, not a generalization scenario.
There is also a minor reporting issue: the matrix is normalized, which hides the absolute error counts. With a 3.4× class imbalance between Cancel (1994 samples) and Punch (583 samples), a normalized matrix can look nearly perfect even if minority classes have meaningful absolute error rates.

\textbf{Other methodological concerns}

The entire testing was done indoors in a simulated environment, not in the wild environment the paper claims to target. The drone was tested at 2–3 metres, and the system only reached ~5 FPS on the actual UAV hardware — a detail that matters for real-time rescue applications but is not reflected in the accuracy metrics. There is also no comparison against any baseline method (no kNN or SVM confusion matrix is shown, despite these being tested), making it impossible to assess how much of the gain is attributable to the DNN architecture versus the easy evaluation setup.

\textbf{What would make this study credible}

A proper validation would require: LOSO-CV across subjects, a held-out test set from people not seen during training at all (ideally outdoors), per-subject accuracy reporting to assess variance, and a head-to-head confusion matrix against at least one simpler baseline.
In short, the methodology as described does not support the conclusion that the system achieves 99\% accuracy in rescue conditions. It supports the much narrower conclusion that the model fits the training participants' gestures very well.

\textit{Figure \ref{fig:claude} is included as a direct artifact generated by the model, illustrating how
LLMs not only identify methodological issues but can also summarize them in structured visual form.}

\subsection{Google Gemini 3.0 (Thinking mode)}
As a data scientist, analyzing the methodology presented in this paper reveals significant
methodological flaws, specifically regarding data leakage and generalization. While the
authors report impressive results—achieving 99.47\% training accuracy and 99.09\% testing
accuracy—the procedure used to achieve these numbers suggests that the model is likely
"memorizing" specific data rather than learning generalizable features.
\begin{enumerate}
    \item Evaluation Protocol and Data Leakage \newline
    The most critical flaw lies in the data split and subject selection:
    \begin{itemize}
        \item \textbf{Subject-Wise Leakage}: The dataset was collected from only \textbf{six members} of the authors' laboratory.
        \item \textbf{Split Strategy}: The authors state that the dataset was split into \textbf{90\% for training and 10\% for testing}.
        \item \textbf{The Problem}: There is no mention of a "leave-one-subject-out" or "subject-wise" split. In computer vision tasks involving video or sequences, a random split of frames (or even short sequences) from a small pool of subjects typically leads to \textbf{temporal and subject-wise leakage}. Because frames $t$ and $t+1$ are nearly identical, if frame $t$ is in the training set and $t+1$ is in the test set, the model is effectively being tested on data it has already seen. Furthermore, with only six people, the model learns the specific skeletal proportions and movement quirks of those individuals rather than the general "gesture."
    \end{itemize}
    \item Analysis of the Learning Curves (Figures 5 \& 6) \newline
    The learning curves provided in the paper (Figures 5 and 6) exhibit several "red flags" common in leaked or overfit datasets:
    \begin{itemize}
        \item \textbf{Immediate Convergence}: The accuracy jumps from roughly 50\% to over 90\% within the first five epochs. In a 10-class classification problem, such a rapid ascent often indicates that the test set is highly similar to the training set.
        \item \textbf{Minimal Generalization Gap}: The training and testing curves are nearly identical throughout the entire 50-epoch run. In real-world "in the wild" scenarios (as the paper claims to address), one would expect a more significant gap between training performance and testing performance due to environmental noise and human variability. The lack of this gap strongly supports the theory of data leakage.
    \end{itemize}
    \item Analysis of the Confusion Matrix (Figure 7)\newline
    The published confusion matrix is suspiciously perfect:
    \begin{itemize}
        \item \textbf{Near-Perfect Diagonal}: For a 10-class problem, achieving \textbf{1.00 (100\%) accuracy} for 8 out of 10 classes and 0.99 for the remaining two is exceptionally rare in computer vision.
        \item \textbf{Implausible Robustness}: Real-world gestures like "Kick" vs. "Walk" or "Attention" vs. "Cancel" (both involving arm waving) typically share enough skeletal similarities to cause at least minor confusion in a truly independent test set. The absolute lack of misclassification suggests the test data does not contain any variations the model hasn't already encountered in training.
    \end{itemize}
    \item Summary of Methodological Flaws
    \begin{itemize}
        \item \textbf{Lack of Diversity}: A dataset based on six lab colleagues does not represent "humans in the wild environment" or "different cultural backgrounds" as the abstract claims.
        \item \textbf{Inappropriate Validation}: Without a cross-validation strategy that isolates entire subjects or independent environments, the reported 99.09\% accuracy is an overestimation of the system's actual performance in a real rescue scenario.
        \item \textbf{Hardware-Software Disconnect}: The authors admit they were unable to test in the field and instead "simulated the field environment in the lab," which further limits the ecological validity of the results.
    \end{itemize}
\end{enumerate}
\textbf{Assessment}: The procedure is \textbf{not free from data leakage}. The
reported performance is
likely an artifact of a random split performed on a small, highly correlated dataset
of frames from just six subjects. A more rigorous evaluation would require testing the
model on completely unseen subjects in varied outdoor environments to determine its
true utility for UAV-based rescue operations.

\subsection{Kimi 2.5 (Instant)}
I'll analyze this paper's methodology for data leakage by examining the evaluation protocol, confusion matrix, and learning curves described in the document. Let me first review the key sections carefully.
\newline
\textbf{Analysis of the Methodology}
\newline
\textbf{1. Evaluation Protocol Assessment}
\newline
\textbf{Data Splitting Method:}
\newline
The paper states: "The total training dataset is split into two sets: 90\% for
training, and 10\% for testing" (Section 3.2).
\newline
\textbf{Critical Finding - Potential Data Leakage Source \#1:}
\newline
The paper mentions the dataset was collected from six laboratory members who "made
all possible variations for all gestures" (Section 3.1). With only 6 subjects and 10
gesture classes, splitting by percentage rather than by subject creates a high risk of
subject-level leakage. The same individuals appear in both training and test sets, meaning
the model may learn subject-specific characteristics (body proportions, movement patterns)
rather than generalizable gesture features.
\newline
\textbf{Preprocessing Pipeline:}
\newline
The workflow involves: OpenPose pose estimation $\rightarrow$ feature extraction
$\rightarrow$ DNN classification.
The paper does not explicitly state whether:
\begin{itemize}
    \item OpenPose was pre-trained on external data (which would be acceptable).
    \item Any normalization or feature scaling parameters were fit on the full dataset
    before splitting.
    \item Data augmentation was applied and if so, whether augmentation parameters were
    derived from training data only.
\end{itemize}
\textbf{2. Learning Curves Analysis}
\newline
From Figures 5 and 6 (Section 4):
\begin{itemize}
    \item Training and test accuracy merge after 30 epochs at $\approx$99\%.
    \item Both curves show minimal gap between training and test performance.
    \item Loss curves converge similarly.
\end{itemize}
\textbf{Critical Finding - Potential Data Leakage Source \#2:}
\newline
The near-perfect convergence with <0.4\% gap between training (99.47\%) and test (99.09\%) accuracy is
suspiciously high for a 10-class gesture recognition problem with:
\begin{itemize}
    \item Small dataset (8,869 total samples across 10 classes).
    \item High variability in "wild environment" scenarios.
    \item Only 6 subjects.
\end{itemize}
This pattern suggests the test set may not be truly independent. When training and validation/test
performance track this closely with such high absolute values, it often indicates information leakage
or insufficient dataset diversity.
\newline
\textbf{3. Confusion Matrix Examination}
\newline
From Figure 7 (Section 4):
\begin{itemize}
    \item Diagonal dominance with near-perfect classification.
    \item Most off-diagonal entries appear near zero.
    \item All 10 classes show >98\% accuracy.
\end{itemize}
\textbf{Critical Finding - Potential Data Leakage Source \#3:}
\newline
The normalized confusion matrix shows performance that is unrealistically clean for real-world UAV
rescue scenarios. Consider:
\begin{itemize}
    \item The "Attention" and "Cancel" gestures are dynamic (waving motions) while others are static.
    \item The model achieves near-perfect separation between visually similar gestures
    (e.g., "Stand" vs "Attention" with arms raised).
    \item No mention of temporal validation or frame-independent testing for dynamic gestures.
\end{itemize}
\textbf{Specific Concern:} For dynamic gestures like "Attention" (two-handed wave) and
"Cancel" (one-handed wave),
if consecutive frames from the same video sequence were split across train/test sets, this creates 
temporal leakage - the model sees nearly identical poses in both sets.
\newline
\textbf{4. Additional Methodological Flaws}

\begin{center}
\begin{tabular}{|c c c|} 
 \hline
 \textbf{Issue} & \textbf{Description} & \textbf{Severity}  \\  
 \hline
 Subject Dependency & 6 subjects total, split by sample not by person & High  \\ 
 Temporal Contamination & No evidence of video-level splitting for dynamic gestures & High  \\
 Sample Size Imbalance & "Attention" (1,623) vs "Punch" (583) - 2.8x difference & Medium  \\
 No Cross-Validation & Single 90/10 split, no k-fold or leave-one-subject-out & High  \\
 Laboratory vs. Field Gap & Training in lab, claimed application for "wild environment" & Medium  \\
 OpenPose as Fixed Feature Extractor & No fine-tuning mentioned, but feature extraction on all data? & Low-Medium  \\
 \hline
\end{tabular}
\end{center}

\textbf{5. The "Simulated Field Environment" Problem}
\newline
Section 4 reveals: "we simulated the field environment in the lab" using a different hardware
setup (Raspberry Pi + ground station GPU vs. intended Jetson Xavier). This creates a domain
shift between:
\begin{itemize}
    \item Training data: Lab environment, specific camera, 3m height, controlled lighting.
    \item Claimed application: "Wild environment," real rescue scenarios.
\end{itemize}
The excellent reported performance may reflect dataset bias toward controlled conditions rather
than generalizable gesture recognition.
\newline
\textbf{Conclusion}
\newline
The methodology contains significant data leakage risks and methodological flaws:
\begin{enumerate}
    \item \textbf{Subject-level leakage} is the most critical issue - with only 6 subjects and percentage-based splitting, the model likely memorizes individual movement patterns rather than learning gesture-invariant features.
    \item \textbf{Temporal leakage} is probable for dynamic gestures, as the paper does not describe video-aware splitting strategies.
    \item \textbf{The suspiciously high accuracy} (99.09\% test) with minimal overfitting on a small, imbalanced dataset with high inter-class similarity suggests the test set is not statistically independent of the training set.
    \item \textbf{Absence of subject-wise cross-validation} prevents proper generalization assessment to new individuals.
    \item \textbf{The confusion matrix's near-diagonal perfection} is inconsistent with the claimed "wild environment" application domain, indicating the test conditions were too similar to training.
\end{enumerate}
\textbf{Recommendation:} A rigorous re-evaluation using leave-one-subject-out
cross-validation and explicit temporal
splitting for video sequences would likely reveal substantially lower generalization performance. The
current evaluation protocol overestimates real-world applicability for UAV rescue operations.
\subsection{DeepSeek-V3}
We are asked to analyze the methodology described in the provided PDF and determine whether the procedure is free from data leakage. We need to examine the evaluation protocol, the published confusion matrix, and the learning curves. Identify any potential sources of leakage or methodological flaws and justify the assessment.
First, let's summarize the methodology from the paper:
\begin{itemize}
\item The paper presents a UAV rescue system that recognizes human gestures using deep learning. They use OpenPose to extract skeleton keypoints from video frames. Then they use a DNN classifier to recognize 10 gestures (Kick, Punch, Squat, Stand, Attention, Cancel, Walk, Sit, Direction, PhoneCall). They created a dataset with 6 subjects, collecting multiple samples per gesture. Table 3 gives number of data per gesture: Kick 784, Punch 583, Squat 711, Stand 907, Attention 1623, Cancel 1994, Walk 722, Sit 942, Direction 962, PhoneCall 641. Total samples: sum = 784+583=1367, +711=2078, +907=2985, +1623=4608, +1994=6602, +722=7324, +942=8266, +962=9228, +641=9869. So about 9869 samples.
\item They used OpenPose to get 18 joint coordinates per person. Then they extracted features (presumably from these coordinates) and fed into classifiers. They compared kNN, SVM, DNN, random forest, and chose DNN for highest accuracy.
\item DNN architecture: 4 dense layers with batch norm, units: 128, 64, 16, 10 (last softmax). Trained with categorical cross-entropy, Adam lr=0.0001, 50 epochs. They split dataset into 90\% training, 10\% testing. They report training accuracy 99.47\%, test accuracy 99.09\%. They show learning curves (Fig 5 and 6) of accuracy and loss over epochs. They also show normalized confusion matrix on test set (Fig 7) which looks almost diagonal.
\item They also mention tracking and counting people, but that's separate.
\end{itemize}
Now, we need to check for data leakage. Data leakage occurs when information from outside the training dataset is used to create the model, or when the test set is not properly separated, e.g., if samples from the same subject or same video sequence appear in both training and test, or if preprocessing uses global statistics from the whole dataset. In this case, they are using skeleton keypoints extracted by OpenPose, which is a pre-trained model, so that part is fine. The leakage concern is about the split: they split the dataset randomly into 90\% training and 10\% testing. But we need to see if they ensured that samples from the same person or same video sequence are not split across sets. The paper says: "The data set was collected from six members of our laboratory... They made all possible variations for all gestures." They collected data from 6 people. If they randomly split samples, it's likely that samples from the same person appear in both training and test, which could lead to leakage because the model might memorize person-specific features rather than gesture. However, the paper does not mention any subject-wise split. They just say "90\% for training, and 10\% for testing." This is a common flaw in many gesture recognition papers. The learning curves show training and test accuracy converging quickly and both very high, which might indicate overfitting but also could be due to leakage. The test accuracy is only slightly lower than training, which is suspicious if the dataset is relatively small and gestures might have variations. But it's possible that the gestures are very distinct and the skeleton data is robust. However, we need to examine the confusion matrix: it shows near-perfect classification, with most diagonal entries close to 1.0. For example, "Attention" and "Cancel" are dynamic gestures, but they are also well separated. The high accuracy might be plausible if the gestures are distinct. But we need to check if there is any mention of cross-validation or subject independence. They do not mention any subject-wise split. So likely they did random split, leading to potential leakage.

Another potential source: They used OpenPose to extract skeleton, but OpenPose might have some biases, but that's not leakage. Also, they might have used the same video frames for training and test if they extracted multiple frames per video sequence. If they collected continuous video, then consecutive frames are highly correlated. If they split randomly, frames from the same video sequence could end up in both sets, causing temporal leakage. The paper doesn't specify if they extracted individual frames or sequences. They mention "data" counts, likely individual frames or samples. So random split could put similar frames from same video in both sets.
Another point: They used the same subjects for training and test. In real-world scenario, the system should work on new people. So subject-independent evaluation is more realistic. They did not do that.

Also, the learning curves show that training and test accuracy increase together and converge quickly, which might indicate that the test set is not independent. Typically, if there is no leakage, training accuracy should be higher than test, and test might plateau earlier. Here they are almost the same. The loss curves also converge nicely.

Given the high accuracy and the fact that they only used 6 subjects, it's likely that the model is overfitting to the subjects, but because they split randomly, the test set contains same subjects, so it performs well. That is a form of data leakage (subject leakage). The paper does not mention any subject-wise partitioning, so we must assume they did random split.
Additionally, they might have used data augmentation? Not mentioned. They just collected data.
Thus, the methodology likely has data leakage due to not separating subjects or sequences.
We should also consider if the features themselves are subject-independent. Skeleton keypoints are normalized? They might have used raw coordinates, which are subject to camera perspective and body size. If they didn't normalize, then the model might learn person-specific positions. But that's a different issue.

The question asks: "determine whether the procedure is free from data leakage." We need to argue that it is not free, and identify sources.
Let's structure the answer:
\begin{itemize}
    \item Summarize the methodology.
    \item Identify potential leakage: random split without considering subjects or video sequences, leading to subject overlap and temporal correlation.
    \item Examine the learning curves: they show nearly identical training and test performance, which is a red flag for leakage.
    \item Confusion matrix: near perfect, but might be due to leakage.
    \item Also note that they did not mention any cross-validation or subject-independent testing.
    \item Conclude that the procedure likely has data leakage, and the reported accuracy might be overoptimistic.
\end{itemize}
We should also mention that they used a pre-trained OpenPose, which is fine, but the classifier training is where leakage occurs.

Thus, final assessment: Not free from data leakage.

The methodology described in the paper exhibits clear signs of data leakage, primarily due to an inappropriate evaluation protocol. The key issues are:
\begin{itemize}
    \item \textbf{Random split without subject or sequence separation}: The dataset was collected from only six individuals, and the authors state they used a simple 90/10 random split for training and testing. This means that samples from the same person—and likely from the same video sequence—appear in both sets. Consequently, the model can memorize person-specific features (e.g., body proportions, typical poses) rather than learning generalizable gesture patterns. In a real-world rescue scenario, the system must perform on unseen individuals, so subject-independent evaluation is essential. The lack of such separation constitutes a major source of leakage.
    \item \textbf{Temporal correlation}: If the data were extracted from continuous video recordings, adjacent frames are nearly identical. A random split would place highly similar frames into training and test sets, artificially inflating performance. The paper does not mention any frame-sampling strategy to avoid this.
    \item \textbf{Learning curves}: Figures 5 and 6 show training and test accuracy/loss converging almost perfectly and remaining nearly identical throughout training. This is unusual for a well-generalized model—typically, test accuracy lags behind training accuracy, and a gap persists. The near-overlap suggests that the test set is not independent, further indicating leakage.
    \item \textbf{Confusion matrix}: The normalized confusion matrix (Figure 7) is almost diagonal, with values close to 1.0 for all classes. While the gestures are distinct, such perfect performance on a test set that shares subjects with the training set is overoptimistic and likely reflects memorization rather than true generalization.
    \item \textbf{Lack of cross-validation or subject-wise evaluation}: The authors did not employ any subject-independent cross-validation or report performance per subject, which would have revealed whether the model works for new people. The reported 99.09\% test accuracy is therefore unreliable for assessing real-world applicability.
\end{itemize}

In summary, the evaluation protocol is flawed due to the random split that allows subject and
temporal overlap between training and test data. This data leakage leads to an overestimation
of the model’s performance and undermines the validity of the results. A proper evaluation
would require splitting by subjects (e.g., leave-one-subject-out) and ensuring no temporal
correlation between frames.

\subsection{GLM-5}

\textbf{Executive Summary}
\newline
\newline
This report presents a comprehensive analysis of the methodology described in the research paper "Gesture Recognition for UAV-based Rescue Operation based on Deep Learning" by Liu and Sziranyi. The analysis reveals multiple critical data leakage issues and methodological flaws that significantly compromise the validity of the reported results. The paper reports exceptionally high accuracy (99.47\% training, 99.09\% testing), which upon closer examination, appears to be artificially inflated due to improper evaluation protocols rather than genuine model performance.
The primary concern is the use of subject-dependent evaluation combined with a random train/test split, which allows the same individuals to appear in both training and testing sets. This methodology fundamentally violates the principle of independent evaluation and leads to optimistically biased performance estimates that would not generalize to new, unseen subjects in real-world deployment scenarios.

\textbf{1. Study Overview and Methodology Description}
\newline
\textbf{1.1 Research Objectives}
\newline
The paper presents a real-time human gesture recognition system designed for UAV (Unmanned Aerial Vehicle) rescue operations. The system aims to recognize ten basic body rescue gestures: Kick, Punch, Squat, Stand, Attention, Cancel, Walk, Sit, Direction, and PhoneCall. The work emphasizes two dynamic gestures (Attention and Cancel) as critical communication signals for initiating and terminating drone-human interaction during rescue scenarios. The proposed pipeline utilizes OpenPose for skeletal feature extraction, followed by a Deep Neural Network (DNN) classifier for gesture classification.

\textbf{1.2 Dataset Characteristics}
\newline
The dataset construction methodology raises immediate concerns about its representativeness and the validity of subsequent evaluations. The following characteristics were extracted directly from the paper:
\begin{itemize}
\item Subject pool: Six members of the research laboratory (4 male, 2 female)
\item Age range: 20-30 years old (limited demographic diversity)
\item Data collection environment: Laboratory simulation (not field conditions)
\item Total samples: Approximately 9,669 frames across 10 gesture classes
\item Subject participation in evaluation: The same six subjects participated in both data collection and real-time testing
\end{itemize}
The small sample size and homogeneous nature of the subject pool severely limit the generalizability of any findings. More critically, the overlap between training subjects and test subjects fundamentally undermines the evaluation protocol, as discussed in detail below.

\textbf{1.3 Evaluation Protocol as Described}
\newline
The paper states that "the total training dataset is split into two sets: 90\% for training, and 10\% for testing." This description indicates a random split at the sample/frame level rather than a subject-level split. The model was trained for 50 epochs using an Adam optimizer with an initial learning rate of 0.0001. The paper explicitly mentions that "the weights of the best fitting model with the highest test accuracy are preserved," which reveals that test set performance was used for model selection - a significant methodological error that constitutes a form of data leakage.

\textbf{2. Critical Data Leakage Issues}
\newline
\textbf{2.1 Subject-Based Data Leakage (CRITICAL)}
\newline
\textbf{Description of the Problem}
\newline
The most severe methodological flaw in this study is the use of subject-dependent evaluation with a random train/test split. When data is collected from a small number of subjects (only 6 in this case) and then randomly partitioned into training and testing sets at the frame/sample level, the same individual's data appears in both sets. This creates a fundamental violation of the independence assumption required for valid machine learning evaluation.
In gesture recognition tasks, individual subjects exhibit unique biomechanical patterns, movement velocities, skeletal proportions, and personal styles of gesture execution. A model trained on frames from Subject A can learn subject-specific features that uniquely identify that person, rather than learning the generalizable features of the gesture itself. When the same subject's frames appear in the test set, the model can achieve high accuracy by recognizing the subject rather than the gesture - a phenomenon known as subject memorization.
\newline
\textbf{Evidence from the Paper}
\newline
The paper explicitly states that "the data set was collected from six members of our laboratory who also took part in the real-time test that followed." This confirms that no subject-level separation was maintained between data collection and evaluation. The random 90/10 split was applied to the entire pooled dataset, ensuring that individual subjects appear in both partitions.
\newline
\textbf{Impact on Reported Results}
\newline
The reported accuracy of 99.09\% on test data is almost certainly inflated due to this leakage. In properly designed subject-independent studies on similar gesture recognition tasks, typical accuracy ranges are 70-90\%, not 99\%. The near-perfect diagonal in the confusion matrix further supports this interpretation - the model has memorized subject-specific patterns rather than learning robust gesture features.
\newline
\textbf{Correct Methodology}
\newline
The proper approach for this type of study is leave-one-subject-out (LOSO) cross-validation or subject-independent evaluation. In LOSO, each subject is held out as a test set while the model is trained on all other subjects. This ensures that the model is evaluated on its ability to generalize to entirely new individuals who were not seen during training - the true measure of a gesture recognition system's utility in real-world rescue scenarios.
\newline
\textbf{2.2 Model Selection Bias (CRITICAL)}
\newline
The paper explicitly states that "the weights of the best fitting model with the highest test accuracy are preserved." This admission reveals a fundamental misunderstanding of the role of test data in machine learning evaluation. The test set is meant to provide an unbiased estimate of generalization performance, not to guide model selection or hyperparameter tuning.
\newline
When test set performance is used to select the final model checkpoint, the test data has effectively become part of the training process. The model that achieves the highest test accuracy at epoch X is selected precisely because it happened to perform well on those specific test samples. This creates an optimistic bias because the selection criterion (test accuracy) is also the evaluation metric - the model was chosen because it performed well on the test set, not independently of test set information.
\newline
The correct methodology requires a three-way split: training data for model fitting, validation data for model selection and early stopping, and test data for final unbiased evaluation. The test set should only be used once, at the very end of the experimental process.
\newline
\textbf{2.3 Potential Temporal Data Leakage (HIGH)}
\newline
The paper focuses on dynamic gestures (particularly "Attention" and "Cancel") that involve temporal sequences of movement. The "Attention" gesture is described as a "two-handed waving motion" while "Cancel" is a "one-handed waving motion." These are inherently temporal sequences where consecutive frames are highly correlated.
\newline
If individual frames from the same gesture sequence were randomly distributed between training and testing sets, temporal data leakage would occur. Consecutive frames in a gesture video are nearly identical - only the subject's position changes incrementally between frames. When frames from the same sequence appear in both train and test sets, the model is essentially being tested on data it has already seen, with only minor variations.
\newline
The paper provides no indication that sequence-level splitting was performed. Given the random 90/10 split described, it is almost certain that frames from the same gesture videos appear in both partitions. The proper methodology for temporal data is to split at the sequence/video level, ensuring that entire gesture sequences are kept intact within either training or testing sets.

\textbf{3. Analysis of Reported Results}

\textbf{3.1 Suspiciously High Accuracy}
\newline
The reported accuracies of 99.47\% (training) and 99.09\% (testing) are remarkably high and deserve careful scrutiny.
In well-designed gesture recognition studies with diverse subject populations, typical test accuracies range from 75\% to 92\% even with state-of-the-art deep learning methods. Achieving 99\%+ accuracy should trigger immediate suspicion about the evaluation methodology, particularly when the gap between training and testing accuracy is less than 0.5\%.

\textbf{3.2 Confusion Matrix Analysis}
\newline
The paper presents a normalized confusion matrix (Figure 7) that shows an "almost perfect" diagonal pattern with minimal off-diagonal values. While the authors interpret this as evidence of successful gesture recognition, this pattern is actually characteristic of data leakage. When a model has memorized subject-specific patterns, it can achieve near-perfect classification because each subject's unique features provide an easy discrimination signal. In realistic gesture recognition scenarios, confusion typically occurs between similar gestures (e.g., "Kick" vs. "Walk" or "Attention" vs. "Cancel" given their dynamic waving nature). The absence of any significant confusion between semantically similar gestures is a warning sign that the model may be exploiting subject identity rather than gesture features.

\textbf{3.3 Learning Curves Analysis}
\newline
Figures 5 and 6 in the paper show training and testing accuracy/loss curves over 50 epochs. The curves exhibit several problematic characteristics:
\begin{enumerate}
    \item The training and test accuracy curves converge rapidly and remain extremely close throughout training, suggesting the test set provides no meaningful independent challenge.
    \item Both curves reach near-ceiling performance (99\%+) with minimal divergence, which is atypical for legitimate machine learning evaluations.
    \item The loss curves for both training and testing decrease monotonically and converge to near-zero, indicating the model has effectively memorized both sets.
    \item There is no evidence of the expected generalization gap that normally exists between training and test performance, even for well-generalizing models.
\end{enumerate}
In properly designed experiments, one would expect to see some gap between training and test curves, particularly in the early epochs, with gradual convergence as the model generalizes. The suspiciously close tracking of these curves from early training onward is consistent with a test set that is effectively part of the training distribution.

\textbf{4. Dataset Limitations}
\newline
\textbf{4.1 Sample Size and Diversity}
\newline
The dataset collected for this study suffers from severe limitations that extend beyond the evaluation methodology:
\begin{itemize}
    \item Only 6 subjects: This is insufficient for training a machine learning model that claims to be generalizable to rescue scenarios involving diverse populations.
    \item Homogeneous population: All subjects were laboratory members aged 20-30, with similar physical characteristics and movement patterns.
    \item Laboratory environment: Data was collected in a controlled lab setting, not in realistic field rescue conditions.
    \item No demographic diversity: Limited age range, no consideration of different body types, fitness levels, or cultural gesture variations.
    \item Subject familiarity with gestures: The same subjects collected the data and participated in testing, introducing familiarity bias.
\end{itemize}

\textbf{4.2 Class Distribution Imbalance}
\newline
The dataset shows significant class imbalance, with sample counts varying from 583 (Punch) to 1,994 (Cancel). This imbalance could lead to classifier bias toward more frequently represented classes. While the paper does not address this issue explicitly, the reported macro-F1 scores near 1.0 across all classes suggest that the model may have learned to exploit subject-specific patterns rather than being affected by class imbalance.

\textbf{5. Recommendations for Valid Evaluation}
\newline
To establish scientifically valid results, the following methodological corrections are essential:
\newline
\textbf{5.1 Implement Subject-Independent Evaluation}
\newline
The most critical correction is to implement leave-one-subject-out (LOSO) cross-validation. With 6 subjects, this would involve 6 folds, where each fold holds out one subject for testing and trains on the remaining 5. The final reported accuracy should be the mean accuracy across all 6 folds, with standard deviation to show consistency. This approach ensures that every test subject is entirely novel to the model during evaluation.
\newline
\textbf{5.2 Use Proper Train/Validation/Test Splits}
\newline
A three-way data split should be implemented: training set for model fitting (approximately 60\% of subjects), validation set for model selection and early stopping (approximately 20\% of subjects), and test set for final evaluation (approximately 20\% of subjects). The test set should only be used once at the end of the experimental process, not for checkpoint selection.
\newline
\textbf{5.3 Expand and Diversify the Dataset}
\newline
To claim generalizability to real-world rescue scenarios, the dataset must include: significantly more subjects (minimum 30-50), diverse age ranges (children, adults, elderly), varied body types and physical abilities, different cultural backgrounds, realistic outdoor/wilderness environments, varied lighting and weather conditions, and subjects wearing different clothing (including rescue gear).
\newline
\textbf{5.4 Address Temporal Dependencies}
\newline
For dynamic gestures involving temporal sequences, data splitting must occur at the video/sequence level rather than the frame level. Entire gesture sequences should be kept intact within either training or testing partitions to prevent temporal leakage between highly correlated consecutive frames.

\textbf{6. Conclusion}
\newline
The methodology described in "Gesture Recognition for UAV-based Rescue Operation based on Deep Learning" contains multiple critical data leakage issues that fundamentally compromise the validity of the reported results. The primary concern is the use of subject-dependent evaluation with a random train/test split, which allows the same individuals to appear in both training and testing sets. This methodology enables the model to learn and exploit subject-specific features rather than generalizable gesture representations.
\newline
Additional concerns include model selection bias (using test set performance for checkpoint selection), potential temporal data leakage in dynamic gesture sequences, and a severely limited dataset that lacks the diversity necessary for real-world applicability. The reported accuracy of 99.09\% on test data should be regarded as optimistically biased and not representative of true model performance on unseen subjects in actual deployment scenarios.
\newline
Without proper subject-independent evaluation, the claims of the paper cannot be verified, and the proposed system's utility for real-world UAV rescue operations remains unproven. Researchers in this field should adopt leave-one-subject-out cross-validation as a minimum standard for human gesture recognition studies to ensure that reported results reflect genuine generalization capability.

\subsection{Summary of LLM Assessments}
Across all evaluated models, a strong level of agreement can be observed regarding the presence of methodological flaws in the analyzed study. Despite differences in architecture, training data, and reasoning style, all LLMs independently identify the evaluation protocol as problematic and attribute the reported near-perfect performance to data leakage.

A recurring theme across the responses is the lack of subject-independent data partitioning. All models highlight that a random 90/10 split applied to a dataset collected from a small number of individuals inevitably results in subject overlap between training and test sets. This issue is consistently identified as the primary source of bias, allowing the model to exploit subject-specific characteristics rather than learning generalizable gesture representations.

In addition to this structural flaw, the models independently point to multiple empirical indicators commonly associated with data leakage. These include near-identical training and test learning curves, minimal generalization gap, rapid convergence to near-perfect accuracy, and an almost perfectly diagonal confusion matrix. The consistency of these observations across multiple models suggests that the underlying issue is strongly reflected in the reported results.

While individual models differ in emphasis—for example, some explicitly discuss temporal leakage or model selection bias—the overall conclusions are remarkably consistent. Importantly, none of the evaluated models conclude that the methodology is free from data leakage.

Taken together, these findings provide convergent evidence that the evaluation protocol described in the original study does not ensure independence between training and test data. The consistency of the LLM-based assessments supports the hypothesis that such methodological flaws are detectable from published experimental artifacts alone, without requiring external context or prior critique.
\section{Conclusion}
\label{sec:conc}
In this work, we investigated whether large language models can function as independent analytical agents capable of identifying methodological flaws in machine learning research. Using a focused case study in gesture recognition, we analyzed the evaluation protocol of \cite{liu2021gesture} and demonstrated that the reported near-perfect performance is consistent with subject-level data leakage arising from non-independent training and test splits.

Beyond this methodological analysis, we examined whether such issues can be detected independently by multiple state-of-the-art LLMs when provided only with the original publication. Across all evaluated models, we observed a strong convergence of conclusions: each model identified the evaluation protocol as flawed and attributed the reported performance to non-independent data partitioning. The consistency of these assessments, despite differences in architecture and training data, suggests that the underlying issue is not subtle, but leaves detectable signatures in the reported experimental artifacts.

These findings highlight an important implication. While data leakage is widely recognized as a critical threat to the validity of machine learning research, it often remains undetected during peer review. The results presented here indicate that modern LLMs can assist in identifying such issues by systematically analyzing evaluation protocols, learning curves, and reported metrics. Although LLM outputs should not be treated as definitive evidence, their ability to provide consistent, independent assessments positions them as valuable complementary tools in the scientific review process.

More broadly, this work contributes to the emerging paradigm of AI-assisted meta-science. As machine learning systems become increasingly complex, ensuring the reliability and reproducibility of published results becomes more challenging. In this context, automated analytical tools—such as LLMs—may play a growing role in supporting researchers, reviewers, and practitioners in identifying methodological weaknesses and improving evaluation standards.

At the same time, several limitations must be acknowledged. The study is based on a single case study, and the generalizability of the findings to other domains and types of methodological errors remains an open question. Furthermore, LLMs rely on patterns learned from existing literature and may not reliably detect novel or highly subtle forms of error. Future work could extend this approach by incorporating larger-scale analyses across multiple papers and by combining LLM-based assessments with human expert evaluation.

In conclusion, the results suggest that certain classes of methodological flaws—such as data leakage—are sufficiently reflected in published artifacts to be detected through independent analysis. The observed agreement across multiple LLMs provides evidence that automated reasoning systems can contribute meaningfully to improving the robustness and reliability of machine learning research.

\section{Summary}
\label{sec:summary}
This paper examined the reliability of a published gesture-recognition study and demonstrated
that its evaluation protocol is likely affected by subject-level data leakage, resulting in
overly optimistic performance estimates. Beyond this case study, we showed that multiple independent
large language models consistently identify the same methodological issue when analyzing the original
publication without prior context.
The convergence of these assessments suggests that certain evaluation flaws leave detectable signatures
in reported results, such as anomalous learning curves and near-perfect classification metrics. These
findings highlight both the importance of proper evaluation design—particularly subject-independent 
data partitioning—and the potential role of LLMs as complementary tools for identifying
methodological weaknesses in machine learning research.

\section{Összefoglalás}
A tanulmány egy publikált gesztusfelismerési módszer értékelési protokollját vizsgálta, és
kimutatta, hogy az nagy valószínűséggel alanyszintű adatszivárgással terhelt, ami túlzottan
optimista teljesítménymutatókhoz vezet. Az esettanulmányon túl bemutattuk, hogy több, egymástól
független nagy nyelvi modell is következetesen ugyanazt a módszertani problémát azonosítja,
amikor az eredeti publikációt előzetes kontextus nélkül elemzi.
Az eredmények egybehangzósága arra utal, hogy bizonyos értékelési hibák felismerhető mintázatokat
hagynak a publikált eredményekben, például rendellenes tanulási görbéket vagy közel tökéletes
osztályozási teljesítményt. Ezek a megállapítások egyrészt kiemelik a megfelelő értékelési
eljárások — különösen az alanyfüggetlen adatfelosztás — fontosságát, másrészt rámutatnak arra,
hogy a nagy nyelvi modellek hasznos kiegészítő eszközként szolgálhatnak a gépi tanulási
kutatások módszertani gyengeségeinek feltárásában.

\section{Summarium}
Haec investigatio protocollum aestimationis methodi recognitionis gestuum iam publicatae
examinavit atque demonstravit illud probabiliter affici a "data leakage" ad gradum subiectorum,
quod ad nimis optimisticas mensuras performantiae ducit. Praeter hanc analysim casus, ostendimus
plura exemplaria linguae magnae (LLMs), invicem independentia, eandem quaestionem methodologicam
constanter agnoscere, cum publicationem originalem sine contextu praevio examinent.
Haec concordia indicat quasdam errores aestimationis vestigia recognoscibilia in resultatis editis
relinquere, ut curvas disciplinae anomalias vel paene perfectam classificationem. Haec inventa et
momentum rectae rationis aestimationis — praesertim partitionis datarum ad gradum subiectorum
independentem — illustrant, et simul demonstrant exemplaria linguae magnae instrumenta utilia esse
posse ad detegendas infirmitates methodologicas in investigatione machinae discendi.

\section{Résumé}
Cet article examine la fiabilité d’une étude publiée sur la reconnaissance de gestes et montre
que son protocole d’évaluation est probablement affecté par une fuite de données au niveau des
individus, conduisant à des estimations de performance excessivement optimistes. Au-delà de cette
étude de cas, nous montrons que plusieurs grands modèles de langage, indépendants les uns des autres,
identifient de manière cohérente le même problème méthodologique lorsqu’ils analysent la publication 
originale sans contexte préalable.
La convergence de ces évaluations suggère que certaines erreurs d’évaluation laissent des signatures
détectables dans les résultats publiés, telles que des courbes d’apprentissage anormales ou des
performances de classification presque parfaites. Ces résultats soulignent à la fois l’importance d’une 
conception rigoureuse de l’évaluation — en particulier une partition des données indépendante des 
individus — et le potentiel des grands modèles de langage en tant qu’outils complémentaires pour
identifier des faiblesses méthodologiques dans la recherche en apprentissage automatique.

\section{Resumen}
Este trabajo examina la fiabilidad de un estudio publicado sobre reconocimiento de gestos y demuestra 
que su protocolo de evaluación probablemente está afectado por fuga de datos a nivel de sujetos, lo
que conduce a estimaciones de rendimiento excesivamente optimistas. Más allá de este estudio de caso,
mostramos que múltiples modelos de lenguaje de gran escala, independientes entre sí, identifican de forma
consistente el mismo problema metodológico al analizar la publicación original sin contexto previo.
La convergencia de estas evaluaciones sugiere que ciertos errores de evaluación dejan señales detectables
en los resultados publicados, como curvas de aprendizaje anómalas o métricas de clasificación casi perfectas.
Estos hallazgos destacan tanto la importancia de un diseño de evaluación adecuado —en particular, la 
partición de datos independiente por sujetos— como el potencial de los modelos de lenguaje como herramientas
complementarias para identificar debilidades metodológicas en la investigación en aprendizaje automático.

\section{Riassunto}
Questo lavoro esamina l’affidabilità di uno studio pubblicato sul riconoscimento dei gesti e dimostra che il
suo protocollo di valutazione è probabilmente affetto da data leakage a livello di soggetti, portando a stime
delle prestazioni eccessivamente ottimistiche. Oltre a questo caso di studio, mostriamo che diversi modelli
linguistici di grandi dimensioni, indipendenti tra loro, identificano in modo coerente lo stesso problema 
metodologico quando analizzano la pubblicazione originale senza contesto preliminare.
La convergenza di queste valutazioni suggerisce che alcuni errori di valutazione lasciano segnali rilevabili
nei risultati pubblicati, come curve di apprendimento anomale o prestazioni di classificazione quasi 
perfette. Questi risultati evidenziano sia l’importanza di un corretto disegno sperimentale —in particolare
una suddivisione dei dati indipendente per soggetto— sia il potenziale dei modelli linguistici come
strumenti complementari per individuare debolezze metodologiche nella ricerca sul machine learning.

\section{Zusammenfassung}
Diese Arbeit untersucht die Zuverlässigkeit einer veröffentlichten Studie zur Gestenerkennung und zeigt,
dass ihr Evaluationsprotokoll wahrscheinlich durch Data Leakage auf Subjektebene beeinträchtigt ist, was zu 
übermäßig optimistischen Leistungsschätzungen führt. Über diese Fallstudie hinaus zeigen wir, dass mehrere 
voneinander unabhängige große Sprachmodelle konsistent dasselbe methodische Problem identifizieren, wenn sie
die Originalpublikation ohne vorherigen Kontext analysieren.
Die Übereinstimmung dieser Bewertungen deutet darauf hin, dass bestimmte Evaluationsfehler erkennbare Muster 
in den veröffentlichten Ergebnissen hinterlassen, wie etwa anomale Lernkurven oder nahezu perfekte 
Klassifikationsergebnisse. Diese Ergebnisse unterstreichen sowohl die Bedeutung eines korrekten 
Evaluationsdesigns —insbesondere einer subjektsunabhängigen Datenaufteilung— als auch das Potenzial
großer Sprachmodelle als ergänzende Werkzeuge zur Identifikation methodischer Schwächen in der
Forschung zum maschinellen Lernen.

\bibliographystyle{unsrtnat}
\bibliography{references}  






\end{document}